%% file: acl2022.tex
\pdfoutput=1
\documentclass[11pt]{article}
\usepackage[final]{acl}
\usepackage{times}
\usepackage{latexsym}
\usepackage[T1]{fontenc}
\usepackage[utf8]{inputenc}
\usepackage{pifont}%
\usepackage{graphicx}
\usepackage{xspace}
\usepackage{booktabs}
\usepackage{multirow}
\usepackage{mathtools}
\usepackage{makecell}
\usepackage[show]{chato-notes}
\usepackage{tcolorbox}
\usepackage{arydshln}
\usepackage{subcaption}
\newcommand{\model}{{\textsc{Contrastive-Probe}}\xspace}

\newcommand{\data}{\texttt{MedLAMA}\xspace}

\newcommand{\cmark}{\ding{51}}%
\newcommand{\xmark}{\ding{55}}%

\newcommand{\mask}{\textsc{[Mask]}}
\newcommand{\rel}[1]{\verb~#1~}
\newcommand{\ent}[1]{\textsl{\color{purple}{#1}}}
\newcommand{\entb}[1]{\textsl{\color{black}{#1}}}
\usepackage{txfonts}
\newcommand{\stitle}[1]{\vspace{0.3ex} \noindent{\bf #1}}

\newcommand\blfootnote[1]{%
  \begingroup
  \renewcommand\thefootnote{}\footnote{#1}%
  \addtocounter{footnote}{-1}%
  \endgroup
}
\makeatletter
\def\adl@drawiv#1#2#3{%
        \hskip.5\tabcolsep
        \xleaders#3{#2.5\@tempdimb #1{1}#2.5\@tempdimb}%
                #2\z@ plus1fil minus1fil\relax
        \hskip.5\tabcolsep}
\newcommand{\cdashlinelr}[1]{%
  \noalign{\vskip\aboverulesep
           \global\let\@dashdrawstore\adl@draw
           \global\let\adl@draw\adl@drawiv}
  \cdashline{#1}
  \noalign{\global\let\adl@draw\@dashdrawstore
           \vskip\belowrulesep}}
\makeatother

\title{Rewire-then-Probe: A Contrastive Recipe for Probing Biomedical Knowledge of Pre-trained Language Models}

\author{Zaiqiao Meng$^{\spadesuit \diamondsuit*}$\ \ \ \  Fangyu Liu$^{\spadesuit*}$\ \ \ \   \textbf{Ehsan Shareghi}$^{\clubsuit\spadesuit}$\\ \textbf{Yixuan Su}$^\spadesuit$\ \ \ \  \textbf{Charlotte Collins}$^\spadesuit$\ \ \ \  \textbf{Nigel Collier}$^\spadesuit$ \\
$^\spadesuit$Language Technology Lab, University of Cambridge \\
$^\diamondsuit$Department of Computing Science, University of Glasgow \\
$^\clubsuit$Department of Data Science and AI, Monash University\\

 \texttt{$^\spadesuit$\{zm324, fl399, ys484, cac74, nhc30\}@cam.ac.uk} \\
 \texttt{$^\clubsuit$ehsan.shareghi@monash.edu}
 }

\begin{document}
\maketitle
\begin{abstract}
\input{sec-abstract}

\end{abstract}
\section{Introduction}
\pagestyle{empty}
\input{sec-introduction}

\section{\data}
\input{sec-data}
\section{Existing Multi-token Knowledge Probing Approaches}
\input{sec-multi-token}
\section{\model: Cloze-style Task as a Self-retrieving Game} \label{contrastive_section}
\input{sec-contrastive}

\section{Experiments}
\input{sec-experiments}
\section{Related Work and Discussion}
\input{sec-relatedwork}

\section{Conclusion}
\input{sec-conclusion}

\section*{Acknowledgements}
Nigel Collier and Zaiqiao Meng kindly acknowledges grant-in-aid support from the UK ESRC for project EPI-AI (ES/T012277/1).
\bibliography{references}
\bibliographystyle{acl_natbib}

\input{sec-appendix}

\end{document}

%% file: sec-abstract.tex
Knowledge probing is crucial for understanding the knowledge transfer mechanism behind the pre-trained language models (PLMs). Despite the growing progress of probing knowledge for PLMs in the general domain, specialised areas such as biomedical domain are vastly under-explored. To facilitate this, we release a well-curated biomedical knowledge probing benchmark, \texttt{\data}, constructed based on the Unified Medical Language System~(UMLS) Metathesaurus. We test a wide spectrum of state-of-the-art PLMs and probing approaches on our benchmark, reaching at most $3\%$ of acc@10. While highlighting various sources of domain-specific challenges that amount to this underwhelming performance, we illustrate that the underlying PLMs have a higher potential for probing tasks. To achieve this, we propose \model, a novel self-supervised contrastive probing approach, that adjusts the underlying PLMs without using any probing data. While \model pushes the acc@10 to $24\%$, the performance gap remains notable. Our human expert evaluation suggests that the probing performance of our \model is under-estimated as UMLS does not comprehensively cover all existing factual knowledge. We hope \data and \model facilitate further developments of more suited probing techniques for this domain.\footnote{The data and code implementation are available at \url{https://github.com/cambridgeltl/medlama}.}
\blfootnote{$^*$Equal contribution. This work was done at the University of Cambridge.}

%% file: sec-introduction.tex
Pre-trained language models (PLMs; \citealt{devlin2019bert,liu2019roberta}) have orchestrated incredible progress on myriads of few- or zero-shot language understanding tasks, by pre-training model parameters in a task-agnostic way and transferring knowledge to specific downstream tasks via fine-tuning~\cite{brown2020language,petroni2020kilt}.

To better understand the underlying knowledge transfer mechanism behind these achievements,
many knowledge probing approaches and benchmark datasets have been proposed~\cite{petroni2019language,jiang2020x,kassner2021multilingual,zhong2021factual}.  
This is typically done by 
formulating knowledge triples as cloze-style queries with the objects being masked (see Table~\ref{tab:examples}) and using the PLM to fill the  single~\cite{petroni2019language} or multiple~\cite{ghazvininejad2019mask} \mask{} token(s) without further fine-tuning. 

\begin{table}[t]
    \centering
     \resizebox{0.48\textwidth}{!}{
    \begin{tabular}{lcl}
    \toprule
        &\textbf{Query} & \textbf{\hspace{1cm}Answer(s)} \\
        \cmidrule[2.5pt](lr){2-2}\cmidrule[2.5pt](lr){3-3}
        \textbf{
          \parbox[t]{2mm}{\multirow{5}{*}{\rotatebox[origin=c]{90}{Hard Queries}}}} 
        &\makecell[l]{\textsl{Riociguat} \\ \hspace{1.5cm}\textbf{has physiologic effect} \mask{}.} & \textsl{Vasodilation}\\
        \cdashlinelr{2-3} 
        &\makecell[l]{\textsl{Entecavir} \\ \hspace{1.5cm}\textbf{may prevent} \mask{}.}  & \textsl{Hepatitis B}\\
        \cdashlinelr{2-3} 
        &\makecell[l]{\textsl{Invasive Papillary Breast Carcinoma} \\ \hspace{1.5cm}\textbf{disease mapped to gene} \mask{}.}  & \textsl{[ERBB2 Gene, CCND1 Gene]}\\
        \hline
        \textbf{
          \parbox[t]{2mm}{\multirow{5}{*}{\rotatebox[origin=c]{90}{Easy Queries}}}}&\makecell[l]{\textsl{Posttraumatic arteriovenous fistula}\\ \hspace{1.5cm}\textbf{is associated morphology of} \mask{}.} & \makecell[l]{\textsl{Traumatic arteriovenous}\\ \textsl{fistula}} \\
         \cdashlinelr{2-3} 
        &\makecell[l]{\textsl{Acute Myeloid Leukemia with Mutated RUNX1}\\  
        \hspace{1.5cm}\textbf{disease mapped to gene} \mask{}.}  & \textsl{RUNX1 Gene} \\
        \cdashlinelr{2-3} 
        &\makecell[l]{\textsl{Magnesium Chloride}\\ \hspace{1.5cm}\textbf{may prevent} \mask{}.} &  \textsl{Magnesium Deficiency} \\
        \bottomrule
    \end{tabular}
    }
    \caption{Example probing queries from \data. \textbf{Bold} font denotes UMLS relation.}
    \label{tab:examples}
    \vspace{-1em}
\end{table}
In parallel, it has been shown that specialised PLMs (e.g., BioBERT; \citealt{lee2020biobert}, BlueBERT; \citealt{peng2019transfer} and PubMedBERT; \citealt{gu2020domain}) substantially improve the performance in several  biomedical tasks~\cite{gu2020domain}.  %
The biomedical domain is an interesting testbed for investigating knowledge probing for its unique challenges (including vocabulary size, multi-token entities), and the practical benefit of potentially disposing the expensive knowledge base construction process. 
However, research on knowledge probing in this domain is largely under-explored.

\begin{table*}[t] %
\small
\centering
\resizebox{0.88\textwidth}{!}{
\begin{tabular}{cll}
\toprule
ID & Relation & Manual Prompt \\
\midrule
1 & \texttt{ disease may have associated disease } & The disease [X] might have the associated disease [Y] . \\
2 & \texttt{ gene product plays role in biological process } & The gene product [X] plays role in biological process [Y] . \\
3 & \texttt{ gene product encoded by gene } & The gene product [X] is encoded by gene [Y] . \\
4 & \texttt{ gene product has associated anatomy } & The gene product [X] has the associated anatomy [Y] . \\
5 & \texttt{ gene associated with disease } & The gene [X] is associatied with disease [Y] . \\
6 & \texttt{ disease has abnormal cell } & [X] has the abnormal cell [Y] . \\
7 & \texttt{ occurs after } & [X] occurs after [Y] . \\
8 & \texttt{ gene product has biochemical function } & [X] has biochemical function [Y] . \\
9 & \texttt{ disease may have molecular abnormality } & The disease [X] may have molecular abnormality [Y] . \\
10 & \texttt{ disease has associated anatomic site } & The disease [X] can stem from the associated anatomic site [Y] . \\
11 & \texttt{ associated morphology of } & [X] is associated morphology of [Y] . \\
12 & \texttt{ disease has normal tissue origin } & The disease [X] stems from the normal tissue [Y] . \\
13 & \texttt{ gene encodes gene product } & The gene [X] encodes gene product [Y] . \\
14 & \texttt{ has physiologic effect } & [X] has physiologic effect of [Y] . \\
15 & \texttt{ may treat } & [X] might treat [Y] . \\
16 & \texttt{ disease mapped to gene } & The disease [X] is mapped to gene [Y] . \\
17 & \texttt{ may prevent } & [X] may be able to prevent [Y] . \\
18 & \texttt{ disease may have finding } & [X] may have [Y] . \\
19 & \texttt{ disease has normal cell origin } & The disease [X] stems from the normal cell [Y] . \\
\bottomrule
\end{tabular}
}
\vspace{-1.5mm}
\caption{The 19 relations and their corresponding manual prompts in \data.}
\label{Table:relation_prompts}
\vspace{-4.5mm}
\end{table*}
To facilitate research in this direction, we present a well-curated biomedical knowledge probing benchmark, \data, that consists of 19 thoroughly selected relations. Each relation contains $1k$ queries ($19k$ queries in total with at most 10 answers each), which are extracted from the large UMLS~\cite{bodenreider2004unified} biomedical knowledge graph and verified by domain experts. We use automatic metrics to identify the hard examples based on the hardness of exposing  answers from their query tokens. See Table~\ref{tab:examples} for a sample of easy and hard examples from \data.

A considerable challenge in probing in biomedical domain is handling multi-token encoding of the answers (e.g. in \data only 2.6\% of the answers are single-token, while in the English set of mLAMA;~\citealt{kassner2021multilingual}, 98\% are single-token), where all existing approaches (i.e., {mask predict}; \citealt{petroni2019language}, {retrieval-based}; \citealt{dufter2021static}, and {generation-based}; \citealt{gao2020making}) 
struggle to be effective.\footnote{Prompt-based probing approaches such as AutoPrompt~\cite{shin2020autoprompt}, SoftPrompt~\cite{qin2021learning}, and OptiPrompt~\cite{zhong2021factual} need additional labelled data for fine-tuning prompts, but we restrict the scope of our investigation to methods that do not require task data.} 
For example, the mask predict approach~\cite{jiang2020x}  which performs well in probing multilingual knowledge achieves less than 1\% accuracy on \data.

To address the aforementioned challenge, we propose a new method, \textbf{\model}, that first adjusts the representation space of the underlying PLMs by using a retrieval-based contrastive learning objective (like `\textsl{rewiring}' the switchboard to the target appliances~\citealt{liu2021mirrorwic}) then retrieves answers based on their representation similarities to the queries.
Notably, our \model does not require using the MLM heads during probing, which avoids the vocabulary bias across different models. Additionally, retrieval-based probe is effective for addressing the multi-token challenge, as it avoids the need to generate multiple tokens from the MLM vocabulary. %
We show that
\model facilitates absolute improvements of up-to {\raise.17ex\hbox{$\scriptstyle\mathtt{\sim}$}}5\% and {\raise.17ex\hbox{$\scriptstyle\mathtt{\sim}$}}21\% on the acc@1 and acc@10 probing performance compared with the existing approaches.

We further highlight that the elicited knowledge by \model is not gained from the additional random sentences, but from the original pre-trained parameters, which echos the previous finding of \citet{liu2021fast,glavas-vulic-2021-supervised,DBLP:journals/corr/abs-2111-04198,DBLP:journals/corr/abs-2202-06417}. Additionally, we demonstrate that different state-of-the-art PLMs and transformer layers are suited for different types of relational knowledge, and different relations requires different depth of tuning, suggesting that both the layers and  tuning depth should be considered when infusing knowledge over different relations. Furthermore, expert evaluation of PLM responses on a subset of \data highlights that expert-crafted resources such as UMLS still do not include the full spectrum of factual knowledge, indicating that the %
factual information encoded in PLMs is richer than what is reflected by the %
automatic evaluation.

The findings of our work, along with the proposed \data and \model, highlight both the unique challenges of the biomedical domain and the unexploited potential of PLMs. We hope our research to shed light on  what  domain-specialised PLMs capture and how it could be better resurfaced, with minimum cost, for probing.

%% file: sec-data.tex
To facilitate research of knowledge probing in the biomedical domain, we create the \data benchmark based on the largest biomedical knowledge graph UMLS~\cite{bodenreider2004unified}. UMLS\footnote{Release version  2021AA: \url{https://download.nlm.nih.gov/umls/kss/2021AA/umls-2021AA-full.zip}} is a comprehensive metathesaurus containing 3.6 million entities and more than 35.2 million knowledge triples over 818 relation types which are integrated from various ontologies, including SNOMED CT, MeSH and the NCBI taxonomy.

\definecolor{darkbrown}{rgb}{0.4, 0.26, 0.13}
Creating a LAMA-style~\cite{petroni2019language} probing benchmark from such a knowledge graph poses its own challenges: (1) UMLS is a collection of knowledge graphs with more than 150 ontologies constructed by different organisations with very different schemata and emphasis; (2) a significant amount of entity names (from certain vocabularies) are unnatural language (e.g., 
\textcolor{darkbrown}{\textsl{t(8;21)(q22;q22)}} denoting an observed karyotypic abnormality) which can hardly be understood by the existing PLMs, with tokenisation tailored for natural language; (3) some queries (constructed from knowledge triples) can have up to hundreds of answers (i.e., 1-to-N relations), complicating the interpretation of probing performance; and (4) some queries may expose answers in themselves (e.g., answer within queries), making it challenging to interpret relative accuracy scores.

\stitle{Selection of Relationship Types.}
In order to obtain high-quality knowledge queries, we conducted multiple rounds of manual filtering on the relation level to exclude uninformative relations or relations that are only important in the ontological context but do not contain interesting semantics as a natural language (e.g, taxonomy and measurement relations). We also excluded relations with insufficient triples/entities. Then, we manually checked the knowledge triples for each relation to filter out those that contain unnatural language entities and ensure that their queries are semantically meaningful.
Additionally, in the cases of 1-to-N relations where there are multiple gold answers for the same query, we constrained all the queries to contain at most 10 gold answers. These steps resulted in 19 relations with each containing 1k randomly sampled knowledge queries.  Table~\ref{Table:relation_prompts} shows the detailed relation names and their corresponding prompts.

\begin{figure}
    \centering
    \includegraphics[width=0.32\textwidth]{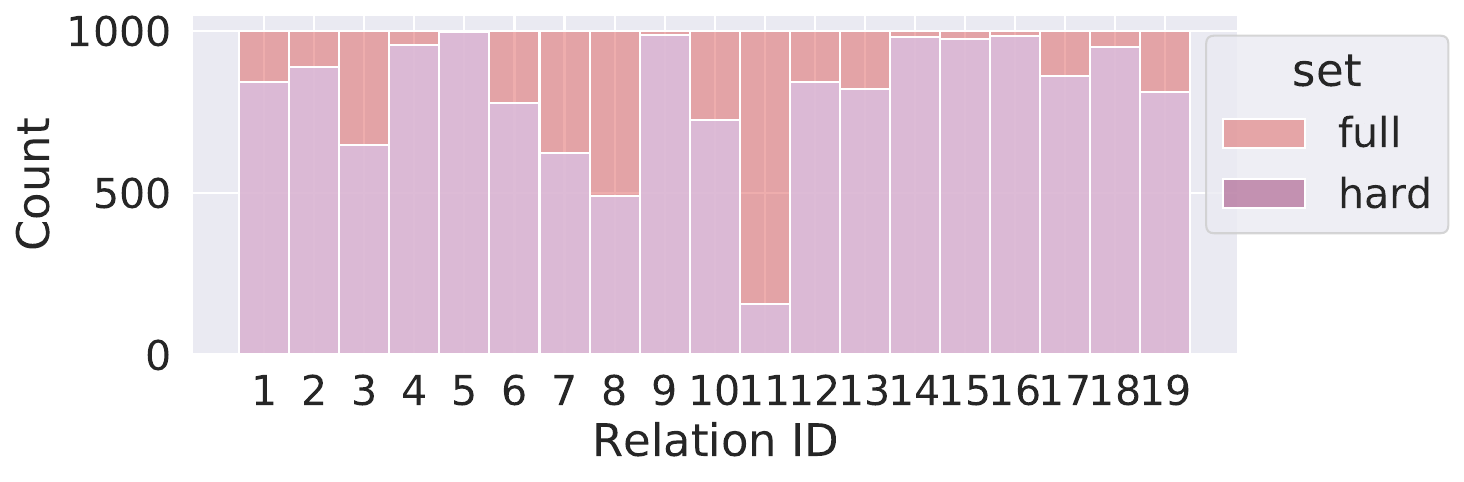}
    \includegraphics[width=0.131\textwidth]{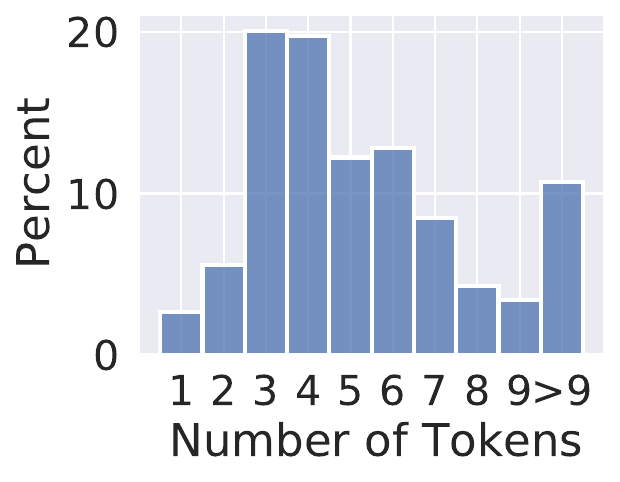}
    \vspace{-0.5em}
    \caption{Left: Count over full and hard sets. Right: Percentage of answers over number of tokens.}
    \label{fig:rel_by_rel}
    \vspace{-1em}
\end{figure}

\stitle{Easy vs. Hard Queries.} Recent works~\cite{poerner2020BERT,shwartz-etal-2020-grounded} have discovered that PLMs are overly reliant on the surface form of entities to guess the correct answer of a knowledge query. The PLMs  ``cheat'' by detecting lexical overlaps between the query and answer surface forms instead of exercising their abilities of predicting factual knowledge. For instance, PLMs can easily deal with the triple
<\textsl{Dengue virus live antigen CYD serotype 1}, \rel{may-prevent}, \textsl{Dengue}> since the answer is part of the query. To mitigate such bias, we also create a hard query set for each relation by selecting a subset of their corresponding 1k queries using token and matching metrics (i.e., exact matching and ROUGE-L~\cite{lin-och-2004-automatic}). For more details see the \emph{Appendix}. We refer to the final filtered and original queries as the \textbf{hard sets} and \textbf{full sets}, respectively. Figure~\ref{fig:rel_by_rel}~(left) shows the count of hard vs. full sets.

\stitle{The Multi-token Issue.} One of the key challenges for probing \data is the multi-token decoding of its entity names. In \data there are only 2.6\% of the entity names that are single-token\footnote{Tokenized by Bert-base-uncased.} while in the English set of mLAMA~\citep{kassner2021multilingual} and LAMA~\cite{petroni2019language} the percentage of single-token answers are  98\% and 100\%, respectively. Figure \ref{fig:rel_by_rel} (right) shows the percentage of answers by different token numbers.
\begin{table}[t]
    \centering
    \resizebox{.48\textwidth}{!}{
    \begin{tabular}{|l|c|c|c|}
    \hline
         \textbf{Approach} &\textbf{Type}& \textbf{Answer space} & \textbf{MLM} \\
         \hline
         Fill-mask~\cite{petroni2019language} & MP &  PLM Vocab & \cmark  \\
         X-FACTR~\cite{jiang2020x} & MP & PLM Vocab  & \cmark \\
         Generative PLMs~\cite{DBLP:conf/acl/LewisLGGMLSZ20} & GB &   PLM Vocab  &\xmark \\
          Mask average~\cite{kassner2021multilingual}& RB & KG Entities &  \cmark   \\
         \model (Ours)  & RB &  KG Entities  & \xmark \\
         \hline
    \end{tabular}
    }
    \caption{Comparison of different approaches. Types of probing approaches: Mask predict (MP), Retrieval-based (RB) and Generation-based (GB).}
    \label{tab:probers}
    \vspace{-1em}
\end{table}

%% file: sec-multi-token.tex
\begin{figure*}[htp]
    \centering
    \vspace{-1em}
    \includegraphics[width=1.0\textwidth]{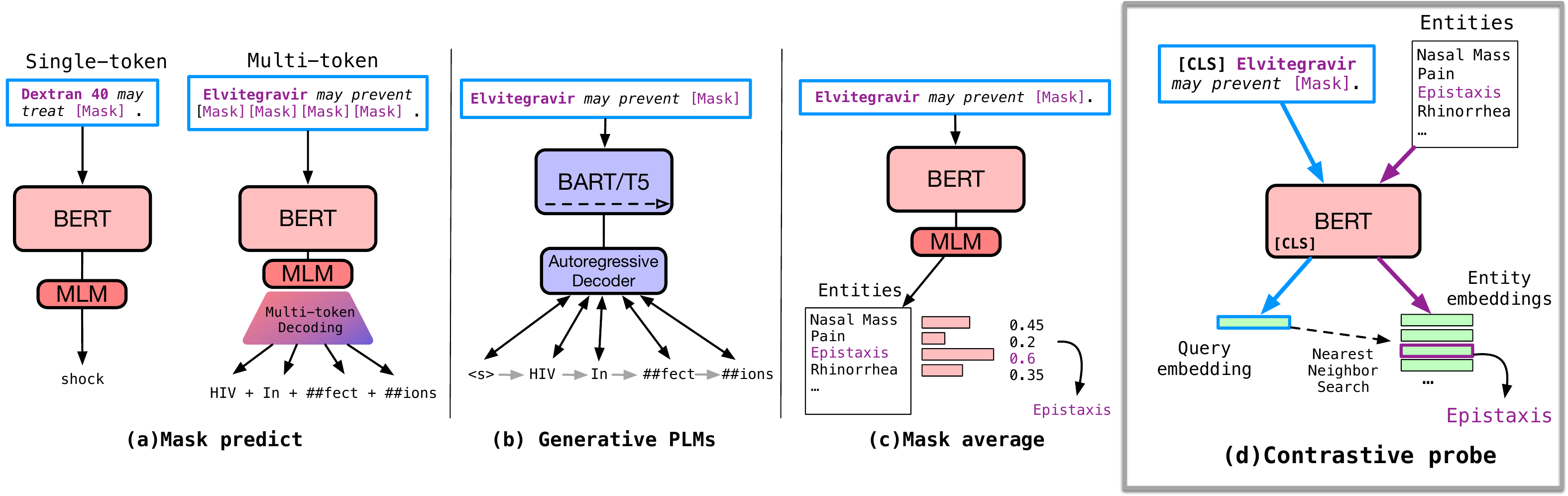}
    \caption{Comparison of different probing approaches. (d) is our proposed \model.}
    \label{fig:probing_approaches}
    \vspace{-1em}
\end{figure*}
While the pioneer works in PLM knowledge probing mainly focused on the single-token entities, many recent works have started exploring the solutions for the multi-token scenario~\cite{kassner2021multilingual,jiang2020x,de2020autoregressive}. These knowledge probing approaches can be categorised, based on answer search space and reliance on MLM head, into three categories: \textit{mask predict}, \textit{generation-based}, and \textit{retrieval-based}. 
Table \ref{tab:probers} summarises their key differences.

\stitle{Mask Predict.}
Mask predict~\cite{petroni2019language,jiang2020x} is one of the most commonly used approaches to probe knowledge for masked PLMs~(e.g. BERT). The mask predict approach uses the MLM head to fill a single mask token for a cloze-style query, and the output token is subjected to the PLM vocabulary~\cite{petroni2019language}. Since many real-world entity names are encoded with multiple tokens, the mask predict approach has also been extended to predict multi-token answers using the conditional masked language model~\cite{jiang2020x,ghazvininejad2019mask}. Figure \ref{fig:probing_approaches}(a) shows the prediction process.
Specifically, given a query, the probing task is formulated as: 1) filling masks in parallel independently (\textit{Independent}); 2) filling masks from left to right autoregressively (\textit{Order}); 3) filling tokens sorted by the maximum confidence greedily~(\textit{Confidence}). After all mask tokens are replaced with the initial predictions, the predictions can be further refined by iteratively modifying one token at a time until convergence or until the maximum number of iterations is reached~\cite{jiang2020x}. For example, \textit{Order+Order} represents that the answers are initially predicted by \textit{Order} and then refined by \textit{Order}. In this paper we examined two of these approaches, i.e. \textit{Independent} and \textit{Order+Order}, based on our initial exploration.

\stitle{Generation-based.}
Recently, many generation based PLMs have been presented for text generation tasks, such as BART~\cite{DBLP:conf/acl/LewisLGGMLSZ20} and T5~\cite{DBLP:journals/jmlr/RaffelSRLNMZLL20}.
These generative PLMs are trained with a de-noising objective to restore its original form autoregressively \cite{DBLP:conf/acl/LewisLGGMLSZ20,DBLP:journals/jmlr/RaffelSRLNMZLL20}. Such an autoregressive generation process is analogous to the \textit{Order} probing approach, thus the generative PLMs can be directly used to generate answers for each query. Specifically, we utilize the cloze-style query with a single \mask{} token as the model input. The model then predicts the answer entities that correspond to the \mask{} token in an autoregressive manner. An illustration is provided in Figure \ref{fig:probing_approaches}(b).

\stitle{Retrieval-based.}
Mask predict and Generation-based approaches need to use the PLM vocabulary as their search spaces for answer tokens, which may generate answers that are not in the answer set. In particular, when probing the masked PLMs using their MLM heads, the predicted result might not be a good indicator for measuring the amount of knowledge captured by these PLMs. This is mainly because the MLM head will be eventually dropped during the downstream task fine-tuning while the MLM head normally accounts for more than 20\% of the total PLM parameters. Alternatively, the retrieval-based probing~\cite{dufter2021static,kassner2021multilingual} are applied to address this issue. Instead of generating answers based on the PLM vocabulary, the retrieval-based approach finds answers by ranking the  knowledge graph candidate entities based on the query and entity representations, or the entity generating scores. To probe PLMs on \data, we use \textit{mask average}~\cite{kassner2021multilingual}, an approach that takes the average log probabilities of entity's individual tokens %
to rank the candidates.
The retrieval-based approaches address the multi-token issue by restricting the output space to the valid answer set and can be used to probe knowledge in different types of PLMs~(e.g. BERT vs. fastText; \citealt{dufter2021static}). However, previous works~\cite{kassner2021multilingual,dufter2021static} only report  results based on the type-restricted candidate set (e.g. relation) which we observed to decay drastically under the full entity set.

%% file: sec-contrastive.tex
To better transform the PLM encoders for the cloze-style probing task, we propose \model which pre-trains on a small number of sentences sampled from the PLM's \emph{original pre-training corpora} with a contrastive self-supervising objective, inspired by the Mirror-BERT~\cite{liu2021fast}. Our contrastive pre-training does not require the MLM head or any additional external knowledge, and can be completed in less than one minute on $2\times2080$Ti GPUs.

\stitle{Self-supervised Contrastive Rewiring.} We randomly sample a small set of sentences (e.g. 10k, see \S\ref{sec:stability} for stability analysis of \model on several randomly sampled sets), and replace their tail tokens (e.g. the last 50\% excluding the full stop) with a \mask{} token. Then these transformed sentences are taken as the queries of the \textit{cloze-style self-retrieving game}. In the following we show an example of transforming a sentence into a cloze-style query:
\begin{tcolorbox}[colback=green!5,colframe=blue!15]
\vspace{-0.5em}
\small
\textbf{Sentence:} Social-distancing largely {\color{purple}{reduces coronavirus infections}}. \\
\textbf{Query:} Social-distancing largely {\color{purple}{\mask{}}}.
\vspace{-0.5em}
\end{tcolorbox}
\noindent where ``{\color{purple}{reduces coronavirus infections}}'' is marked as a positive answer of this query. 

Given a batch, the \textit{cloze-style self-retrieving game} is to ask the PLMs to retrieve the positive answer from all the queries and answers in the same batch. Our \model tackles this by optimising an InfoNCE objective~\cite{oord2018representation},
\begin{equation}
\setlength{\abovedisplayskip}{3pt}
\setlength{\belowdisplayskip}{3pt}
\setlength{\abovedisplayshortskip}{3pt}
\setlength{\belowdisplayshortskip}{3pt}
    \mathcal{L} = -\sum_{i=1}^{N}\log\frac{\exp(\cos(f(x_i), f(x_p))/\tau)}{\displaystyle \sum_{x_j\in \mathcal{N}_i}\exp(\cos(f(x_i), f(x_j))/\tau)},
    \label{eq:infonce}
\end{equation}
where $f(\cdot)$ is the PLM encoder (with the MLM head chopped-off and \texttt{[CLS]} as the contextual representation), $N$ is batch size, $x_i$ and $x_p$ are from a query-answer pair (i.e., $x_i$ and $x_p$ are from the same sentence), $\mathcal{N}_i$ contains queries and answers in the batch, and $\tau$ is the temperature. This objective function encourages $f$ to create similar representations for any query-answer pairs from the same sentence and dissimilar representations for queries/answers belonging to different sentences. 

\stitle{Retrieval-based Probing.} For probing step, the query is created based on the prompt-based template for each knowledge triple
, as shown in the following:
\begin{tcolorbox}[colback=green!5,colframe=blue!15]
\vspace{-0.65em}
\small
\textbf{Triple:} <\textsl{Elvitegravir}, \texttt{may-prevent}, \textsl{\color{purple}{Epistaxis}}>\\
\textbf{Query:} Elvitegravir may prevent {\color{purple}{\mask{}}}.
\vspace{-0.65em}
\end{tcolorbox}
\noindent and we search for nearest neighbours from all the entity representations encoded by the same model.

%% file: sec-experiments.tex
In this section we conduct extensive experiments to verify whether \model is effective for probing biomedical PLMs. First, we experiment with \model and existing probing approaches on \data benchmark~(\S \ref{sec:benchmarking}). Then, we conduct in-depth analysis of the stability and applicability of \model in probing biomedical PLMs~(\S \ref{sec:stability}). Finally, we report an evaluation of a biomedical expert on the probing predictions and highlight our findings (\S \ref{sec:human_evaluation}). 

\stitle{\model Rewiring.} We train our \model based on 10k sentences which are randomly sampled from the PubMed texts\footnote{We sampled the sentences from a PubMed corpus used in the pre-training of BlueBERT~\cite{peng2019transfer}.} using a mask ratio of 0.5. The best hyperparameters and their tuning options are provided in \emph{Appendix}.

\begin{table}[t]
    \centering
    \resizebox{.45\textwidth}{!}{
    \begin{tabular}{llcc}
    \toprule
     	\multirow{2}{*}{\textbf{Approach}} & 	\multirow{2}{*}{\textbf{PLM}} & \multicolumn{2}{c}{\textbf{Full Set}} \\
     	\cmidrule(l){3-4}
     	& & \textbf{acc@1} & \textbf{acc@10}  \\
         \hline
          \multirow{5}{*}{\textbf{Generative PLMs}} & BART-base & 0.16 &  1.39  \\
           & SciFive-base & 0.53 &  2.02  \\
           & SciFive-large & 0.55 &  2.03   \\
           & T5-small & 0.70 &  1.72   \\
           & T5-base & 0.06 &  0.19    \\
         \hline
         \multirow{3}{*}{\textbf{X-FACTR} (\textit{Confidence})} &  BERT & 0.05 & -  \\
         &  BlueBERT& 0.74 & -   \\
         &  BioBERT & 0.17 & -  \\
         \cmidrule(l){1-4}
         \multirow{3}{*}{\textbf{X-FACTR} (\textit{Order}+\textit{Order})} &  BERT & 0.06 & -  \\
         &  BlueBERT & 0.50 & -  \\
         &  BioBERT & 0.11 & - \\
         \hline
         \multirow{3}{*}{\textbf{Mask average}} &   BERT & 0.06 & 0.73 \\
         &  BlueBERT & 0.05 & 1.39  \\
         &  BioBERT & 0.28 & 3.03 \\
         \cmidrule(l){1-4}
         \multirow{4}{*}{\textbf{\model (Ours)}} 
        &  BERT & 1.95 & 6.96 \\
         &  BlueBERT & \underline{4.87} & \underline{19.87} \\
         &  BioBERT & 3.28 & 15.46 \\
         &  PubMedBERT & \textbf{5.71} & \textbf{24.31}  \\
         \bottomrule
    \end{tabular}
    }
    \caption{Performance of different probing approaches on the full set of \data. Since the MLM head of PubMedBERT is not available, the mask predict and mask average approaches cannot be applied. Best results are in \textbf{bold} and the second bests are \underline{underlined}.}
    \label{tab:overall_result}
    \vspace{-1em}
\end{table}
\stitle{Probing Baselines.} For the mask predict approach, we use the original implementation of X-FACTR~\cite{jiang2020x}, and set the beam size and the number of masks to 5. Both mask predict and retrieval-based approaches are tested under both the general domain and biomedical domain BERT models, i.e. Bert-based-uncased~\cite{devlin2019bert}, BlueBERT~\cite{peng2019transfer}, BioBERT~\cite{lee2020biobert}, PubMedBERT~\cite{gu2020domain}.\footnote{The MLM head of PubMedBERT is not publicly available and cannot be evaluated by X-FACTR and \textit{mask average}.}  For generation-based baselines, we test five PLMs, namely BART-base~\cite{DBLP:conf/acl/LewisLGGMLSZ20}, T5-small and T5-base~\cite{DBLP:journals/jmlr/RaffelSRLNMZLL20}
that are general domain generation PLMs, and SciFive-base \& SciFive-large~\cite{phan2021scifive} that are pre-trained on large biomedical corpora.

\subsection{Benchmarking on \data}
\label{sec:benchmarking}
\stitle{Comparing Various Probing Approaches.} Table \ref{tab:overall_result} shows the overall results of various probing baselines on \data. It can be seen that the performances of all the existing probing approaches (i.e. \textit{generative PLMs}, \textit{X-FACTR} and \textit{mask predict}) are very low (<1\% for acc@1 and <4\% for acc@10) regardless of the underlying PLM, which are not effective indicators for measuring knowledge captured.
In contrast, our \model obtains absolute improvements by up-to $\sim5\%$ and $\sim21\%$ on acc@1 and acc10 respectively comparing with the three existing approaches, which validates its effectiveness on measuring the knowledge probing performance. In particular, PubMedBERT model obtains the best probing performance (5.71\% in accuracy) for these biomedical queries, validating its effectiveness of capturing biomedical knowledge comparing with other PLMs (i.e. BERT, BlueBERT and BioBERT).

\stitle{Benchmarking with \model.}
To further examine the effectiveness of PLMs in capturing biomedical knowledge, we benchmarked several state-of-the-art biomedical PLMs (including pure pre-trained and knowledge-enhanced models) on \data through our \model. 
Table \ref{tab:benchmarking} shows the probing results over the full and hard sets. In general, we can observe that these biomedical PLMs always perform better than general-domain PLMs (i.e., BERT). Also, we observe the decay of performance of all these models on the more challenging hard set queries.
While PubMedBERT performs the best among all the pure pre-trained models, SapBERT~\cite{liu2020self} and CoderBERT~\cite{yuan2020coder}~(which are the knowledge infused PubMedBERT) further push performance to 8\% and 30.41\% on acc@1 and acc@10 metrics respectively, highlighting the benefits of knowledge infusion pre-training.

\begin{table}[t]
    \centering
    \resizebox{.49\textwidth}{!}{
    \begin{tabular}{lcc}
    \toprule
     	\multirow{2}{*}{\textbf{Model}} &
     	\multicolumn{2}{c}{\textbf{acc@1/acc@10}} \\
     	\cmidrule[1pt](l){2-3}
     	& \textbf{Full Set} & \textbf{Hard Set} \\
     	\midrule
         BERT~\cite{devlin2019bert} & 1.95$\pm 0.40$/6.96 $\pm 0.96$ & 0.67$\pm 0.19$/3.27$\pm 0.54$\\
         BlueBERT~\cite{peng2019transfer} & 4.87$\pm 0.43$/19.87$\pm 0.62$ & 4.12$\pm 0.46$/18.18$\pm 0.77$ \\  
         BioBERT~\cite{lee2020biobert} & 3.28$\pm 0.20$/15.46$\pm 0.93$ & 2.14$\pm 0.23$/12.59$\pm 1.19$\\
         ClinicalBERT~\cite{alsentzer2019publicly} & 1.83$\pm 0.15$/8.64$\pm 0.79$ & 0.71$\pm 0.13$/5.45$\pm 1.06$\\ %
         SciBERT~\cite{beltagy2019scibert} & 3.64$\pm 0.33$/18.11$\pm 1.95$ & 2.14$\pm 0.30$/14.64$\pm 2.01$\\ %
         PubMedBERT~\cite{gu2020domain} & {5.71$\pm 0.58$}/{24.31$\pm 1.29$} & {4.49$\pm 0.49$}/{21.74$\pm 1.21$} \\
         \cmidrule[1.5pt](l){1-3}
         UmlsBERT~\cite{michalopoulos2021umlsbert} & 2.94$\pm 0.21$/11.64$\pm 0.46$ & 1.80$\pm 0.11$/7.75$\pm 0.42$ \\ 
         SapBERT~\cite{liu2020self} & \underline{7.80$\pm 0.38$}/\textbf{30.41$\pm 1.23$}& \underline{5.15$\pm 0.27$}/\textbf{26.09$\pm 1.17$} \\ %
         CoderBERT~\cite{yuan2020coder} & \textbf{8.00$\pm 0.60$}/\underline{26.41$\pm 1.08$} & \textbf{6.08$\pm 0.52$}/\underline{22.69$\pm 1.10$}\\ %
         \bottomrule
    \end{tabular}
    }
    \caption{Benchmarking biomedical PLMs on \data~(Full and Hard) via \model. The bottom panel are knowledge-enhanced PLMs. The average performance and their standard deviation are reported based on rewiring over 10 different random sets.}
    \label{tab:benchmarking}
    \vspace{-0.5em}
\end{table}

\stitle{Comparison per Answer Length.}
\begin{figure}[t]
    \centering
    \includegraphics[width=0.48\textwidth]{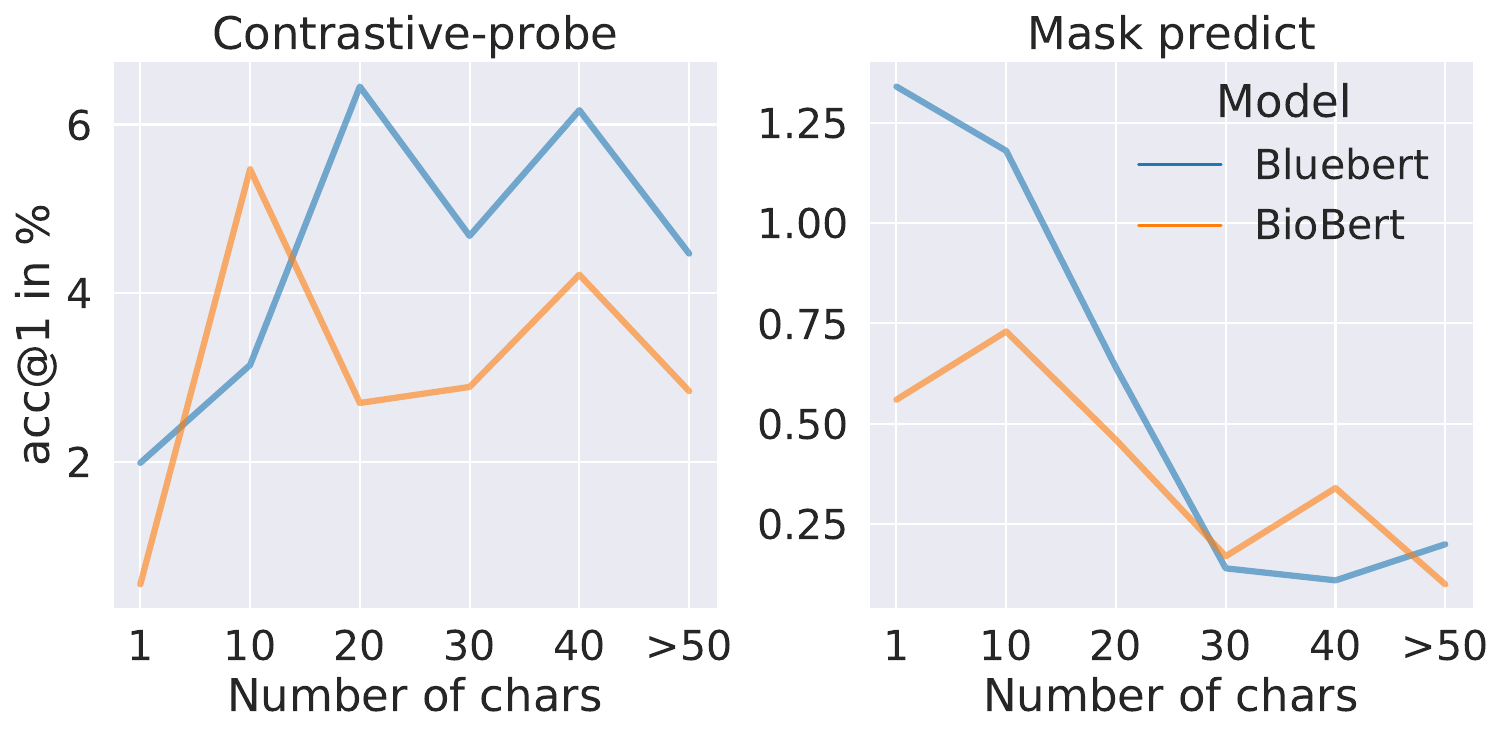}
    \vspace{-2em}
    \caption{Performance over answer lengths.}
    \label{fig:acc_by_char_len}
    \vspace{-1em}
\end{figure}
Since different PLMs use different tokenizers, we use char length of the query answers to split \data into different bins and test the probing performance over various answer lengths. Figure \ref{fig:acc_by_char_len} shows the result. We can see that the performance of retrieval-based probing in \model increases as the answer length increase while the performance of mask predict dropped significantly. 
This result validates that our \model (retrieval-based) are more reliable at predicting longer answers than the mask predict approach since the latter heavily relies on the MLM head.\footnote{For the single-token answer probing scenario, \model does not outperform the mask predict approach, particularly in the general domain. This is expected since most of the masked PLMs are pre-trained by a single-token-filling objective.}

\subsection{In-depth Analysis of \model}
\label{sec:stability}

\begin{figure}[t]
    \centering
     \includegraphics[width=0.48\textwidth]{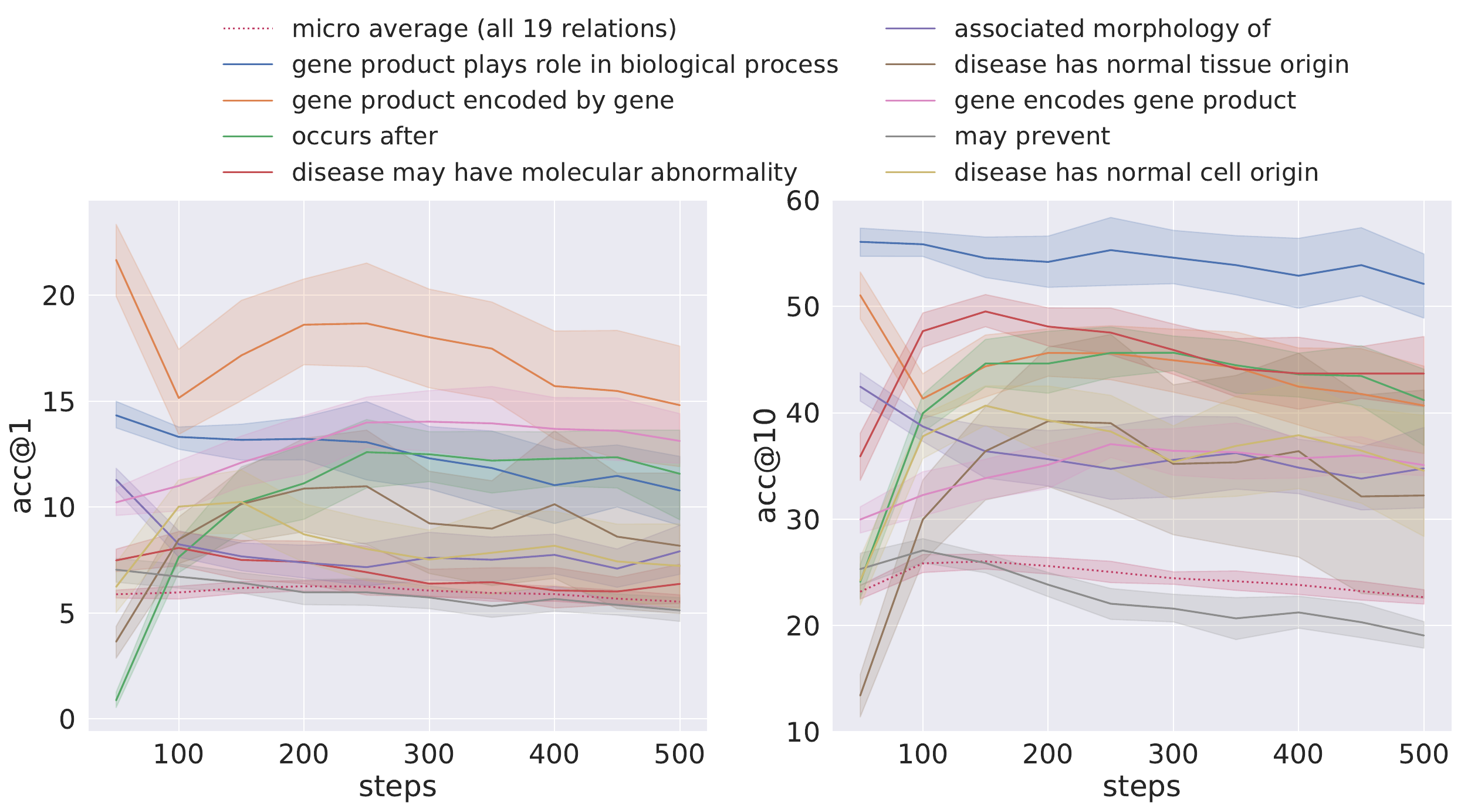}
    \vspace{-2em}
    \caption{Performance over training steps on full set. The shaded regions are the standard deviations.}
    \label{fig:by_relation_step}
    \vspace{-1em}
\end{figure}

Since our \model involves many hyperparameters and stochastic factors during self-retrieving pre-training, it is critical to verify if it behaves consistently under (1) different randomly sampled sentence sets; (2) different types of relations; and (3) different pre-training steps.

\stitle{Stability of \model.} To conduct this verification, we sampled 10 different sets of 10k sentences from the PubMed corpus\footnote{The tuning corpus itself is unimportant, since we can obtain the similar results even using Wikipedia.} and probed the PubMedBERT model using our \model on the full set. 
Figure~\ref{fig:by_relation_step} shows the acc@1 performance over top 9 relations and the micro average performance of all the 19 relations. We can see that the standard deviations are small and the performance over different sets of samples shows the similar trend. This further highlights that the probing success of \model is not due to the selected pre-training sentences. Intuitively, the contrastive self-retrieving game (\S\ref{contrastive_section}) is equivalent to the formulation of the cloze-style filling task, hence tuning the underlying PLMs makes them better suited for knowledge elicitation needed during probing (like `rewiring’ the switchboards).
Additionally, from Figure \ref{fig:by_relation_step} we can also observe that different relations exhibit very different trends during pre-training steps of \model and peak under different steps, suggesting that we need to treat different types of relational knowledge with different tuning depths when infusing knowledge. We leave further exploration of this to future work.

\begin{figure*}
\centering
  \includegraphics[width=0.88\textwidth]{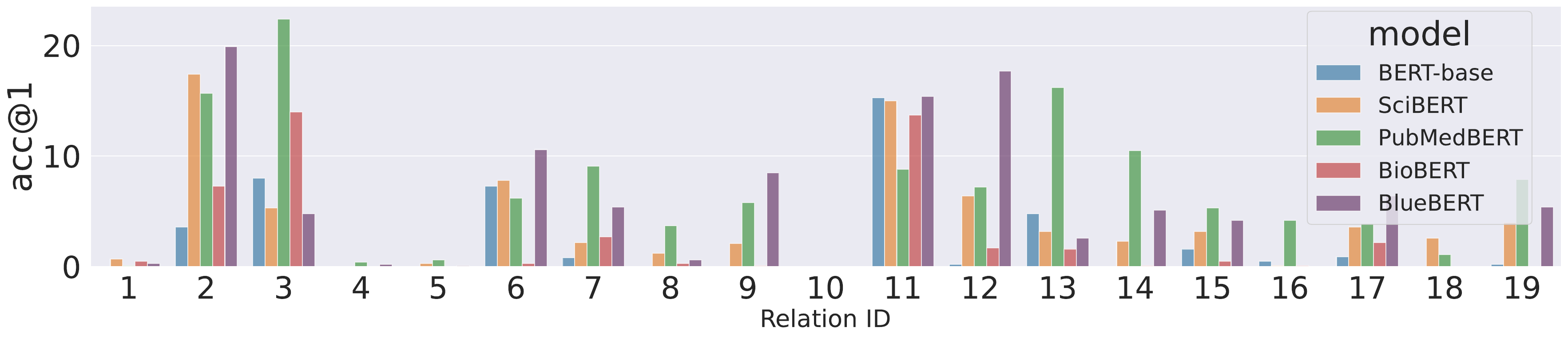}
\vspace{-1em}
\caption{Performance of PLMs over different relations.}
\label{fig:model_by_rel} 
\vspace{-1em}
\end{figure*}

\begin{figure}
    \centering
    \includegraphics[width=0.23\textwidth]{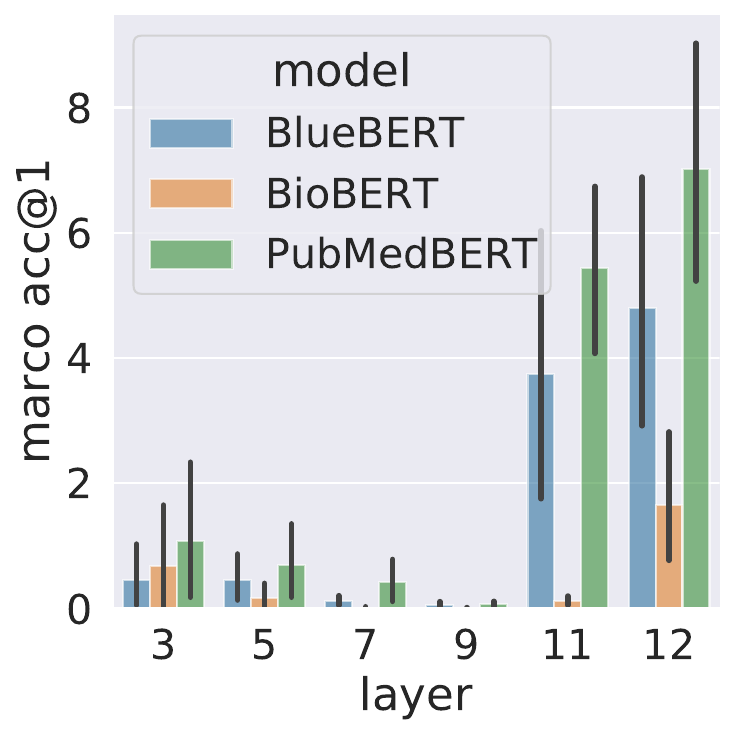}
    \includegraphics[width=0.23\textwidth]{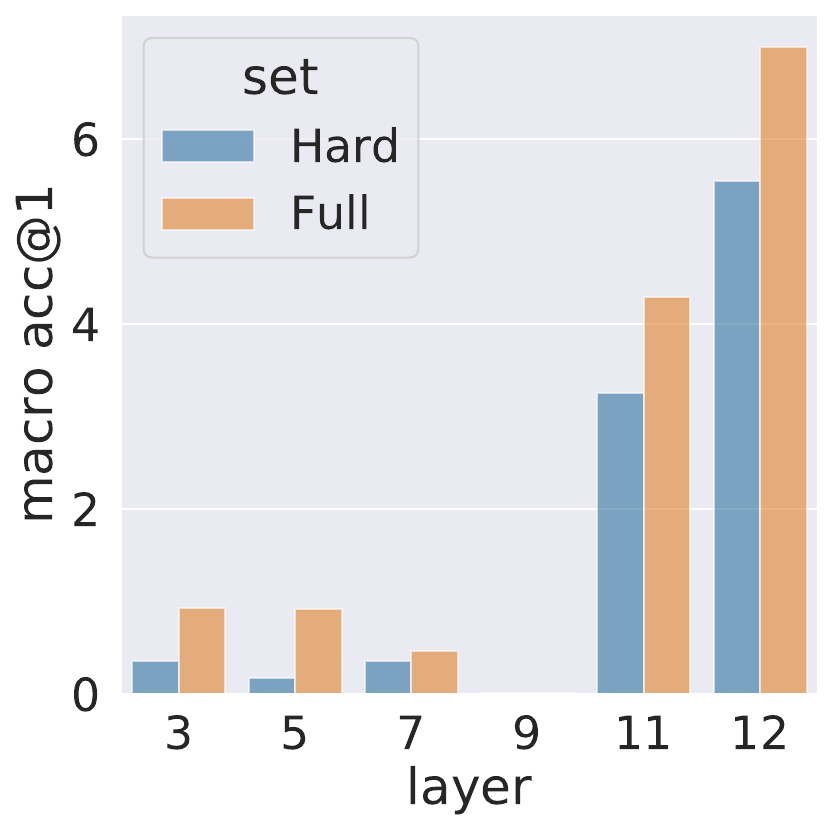}
    \vspace{-1em}
    \caption{Performance over different layers.}
    \label{fig:marco_acc_by_layer}
    \vspace{-1em}
\end{figure}

\stitle{Probing by Relations.} To further analyse the probing  variance over different relations, we also plot the probing performance of various PLMs over different relations of \data in Figure~\ref{fig:model_by_rel}. We can observe that different PLMs exhibit different performance rankings over different types of relational knowledge (e.g. BlueBERT peaks at \textit{relation 12} while PubMedBERT peaks at \textit{relation 3}). This result demonstrates that different PLMs are suited for different types of relational knowledge. We speculate this to be reflective of their training corpora.

\begin{figure*}
\centering
\includegraphics[width=0.94\textwidth]{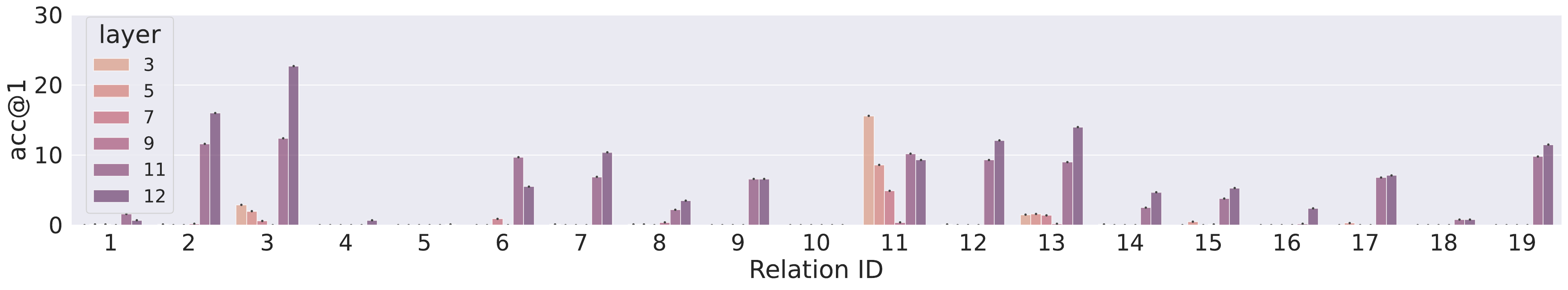}
\vspace{-1em}
\caption{Performance of PubMedBERT over layers.}
\vspace{-1em}
\label{fig:layers_by_rel}
\end{figure*}

\stitle{Probing by Layer.}
To investigate how much knowledge is stored in each Transformer layer, we chopped the last layers of PLMs and applied \model to evaluate the probing performance based on the first $L\in\{3,5,7,9,11,12\}$ layers on \data.
In general, we can see in Figure \ref{fig:marco_acc_by_layer} that the model performance drops significantly after chopping the last 3 layers, while its accuracy is still high when dropping only last one layer. 
In Figure \ref{fig:layers_by_rel}, we further plot the layer-wise probing performance of PubMedBERT over different relations. Surprisingly, we find that different relations do not show the same probing performance trends over layers. For example, with only the first 3 layers, PubMedBERT achieves the best accuracy (>15\%) on \textit{relation 11}
queries. This result demonstrates that both relation types and PLM layers are confounding variables in capturing factual knowledge, which helps to explain the difference of training steps over relations in Figure~\ref{fig:by_relation_step}. This result also suggests that layer-wise and relation-wise training could be the key to effectively infuse factual knowledge for PLMs.

\subsection{Expert Evaluation on Predictions}
\label{sec:human_evaluation}

To assess whether the actual probing performance could be possibly higher than what is reflected by the commonly used automatic evaluation, we conducted a human evaluation on the prediction result.
Specifically, we sample 15 queries and predict their top-10 answers using \model based on PubMedBERT and ask the assessor\footnote{A senior Ph.D. graduate in Cell Biology.} to rate the predictions on a scale of [1,5].
Figure \ref{fig:human_annotation} shows the confusion matrices.\footnote{In the \emph{Appendix}, we provide examples with their UMLS gold answers, human annotated answers and probing predictions of different probing approaches.} We observe the followings: (1) There are 3 UMLS answers that are annotated with score level 1-4 (precisely, level 3), which indicates UMLS answers might not always be the perfect answers. (2) There are 20 annotated perfect answers (score 5) in the top 10 predictions that are not marked as the gold answers in the UMLS, which suggests the UMLS does not include all the expected gold knowledge. (3) In general, PubMedBERT achieves an 8.67\% (13/150) acc@10 under gold answers, but under the expert annotation the acc@10 is 22\% (33/150), which means the probing performance is higher than what evaluated using the automatically extracted answers. 
\begin{figure}
\centering
\includegraphics[trim={0 1cm 0 0}, width=0.4\textwidth]{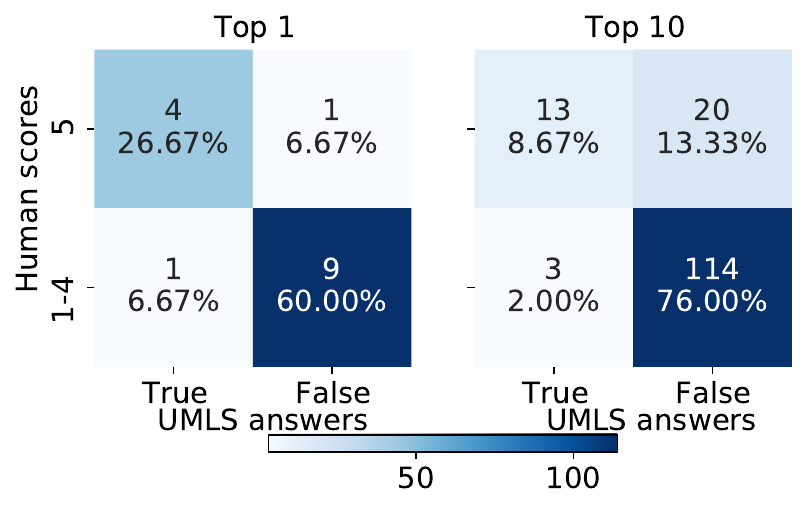}
\caption{Confusion matrices of expert annotated scores versus the extracted UMLS answers. Five annotation score levels: 5-\textit{Perfectly answer the query}; 4-\textit{Similar to the gold answer, could somehow be the answer}; 3-\textit{Related to the query but not correct}; 2-\textit{Same domain or slight relation}; 1-\textit{Completely unrelated}.}
\label{fig:human_annotation}
\vspace{-2.5mm}
\end{figure}

\begin{table}[!ht]
    \centering
    \resizebox{.48\textwidth}{!}{
    \begin{tabular}{cccccccc}
    \toprule
    Benchmark & \# Rel. & \# Queries & Avg. \# Answer & \% Single-Tokens\\
    \midrule
    LAMA & 41 & 41k & 1 & 100\%\\
    BioLAMA & 36 & 49k & 1 & 2.2\% \\
    \data & 19 & 19k & 2.3 & 2.6\%\\
    \bottomrule  
    \end{tabular}
    }
    \vspace{-0.5em}
    \caption{Statistics comparison among LAMA, BioLAMA and our \data.}
    \label{tab:statistics}
\end{table}
\subsection{Comparing with BioLAMA} 
During the writing of this work, we noticed a concurrent work 
to ours that also released a biomedical knowledge probing benchmark, called BioLAMA~\citet{sung2021language}. In Table \ref{tab:statistics}, we compare \data with LAMA~\cite{petroni2019language} and BioLAMA in terms of data statistics. We found that there is only 1 overlapped relation (i.e., may treat) between BioLAMA and \data, and no overlap exists on the queries. 
We can see that, without additional training data from the biomedical knowledge facts, \model reaches  a promising performance compared with OptiPrompt approach, which needs further training data. Additionally, since Mask Predict and OptiPrompt require using the MLM head, it is impossible to compare a model without MLM head being released (e.g. PubMedBERT). In contrast, our \model not only provides a good indicator of comparing these models in terms of their captured knowledge, but also makes layer-wise knowledge probing possible.

\subsection{Limitations of \model}

\stitle{How to early stop?} For fair comparison of different PLMs, we currently use checkpoints after contrastive tuning for a fixed number of steps (200, specifically). However, we have noticed that different models and different probing datasets have different optimal training steps. To truly `rewire' the most knowledge out of each PLMs, we need a unified validation set for checkpoint selection. What the validation set should be and how to guarantee its fairness require further investigation.

\stitle{Performance not very stable.} We have noticed that using different contrastive tuning corpus as well as different random seeds can lead to a certain variance of their probing performances (see Table \ref{tab:benchmarking}). To mitigate such issue, we use average performance of 10 runs on 10 randomly sampled corpus. Improving the stability of \model and investigating its nature is a future challenge.

\begin{table}[t]
    \centering
    \resizebox{.48\textwidth}{!}{
    \begin{tabular}{cccccccc}
    \toprule
     \multirow{2}{*}{\textbf{Probe}} & \multirow{2}{*}{\textbf{Model}} & \multicolumn{2}{c}{\textbf{CTD}} & \multicolumn{2}{c}{\textbf{wikidata}} & \multicolumn{2}{c}{\textbf{UMLS}} \\
 	 \cmidrule(l){3-4} \cmidrule(l){5-6} \cmidrule(l){7-8}
     &  &	acc@1 & acc@5 & acc@1 & acc@5 & acc@1 & acc@5 \\
      \midrule
      \multirow{2}{*}{Mask Predict} & BERT & 0.06 & 1.20 & 1.16 & 6.04  & 0.82 & 1.99  \\
      & BioBERT & 0.42 & 3.25 & 3.67 & 11.20  & 1.16 & 3.82  \\
      & Bio-LM & 1.17 & 7.30 &  \textbf{11.97} & 25.92  &  3.44 & 8.88  \\
      \midrule
      \multirow{2}{*}{OptiPrompt} & BERT & \underline{3.56} & 6.97 & 3.29 & 8.13  & 1.44 & 3.65  \\
      & BioBERT & \textbf{4.82} & \underline{9.74} & 4.21 & 12.91  & \underline{5.08} & 13.28  \\
      & Bio-LM & 2.99 & \textbf{10.19} &  \underline{10.60} & 25.15  &  \textbf{8.25}& \textbf{20.19} \\
      \midrule
      \multirow{3}{*}{\model} & BlueBERT & 1.62 & 5.84 & 6.64 & \underline{25.97} & 2.63 & 11.46 \\
      & BioBERT & 0.20 & 0.99 & 1.04 & 4.51 & 0.89 & 3.89\\
      & Bio-LM & 1.70 & 4.26 & 4.32 & 18.74 & 1.27 & 5.01\\
      & PubMedBERT & 2.60 & 8.87 & 10.20 & \textbf{35.14} & 4.93 & \underline{18.33}\\
      \bottomrule  
    \end{tabular}
    }
    \vspace{-0.5em}
    \caption{Performance on BioLAMA benchmark. Note that both the mask predict and opti-prompt require using the MLM head and opti-prompt needs further training data, so it is impossible to compare a model without MLM head being released (e.g. PubMedBERT). In contrast, our \model make all these models comparable in terms of their captured knowledge.}
    \vspace{-1em}
    \label{tab:biolama}
\end{table}

%% file: sec-relatedwork.tex
\stitle{Knowledge Probing Benchmarks for PLMs.} LAMA \cite{petroni2019language}, which starts this line of work, is a collection of single-token knowledge triples extracted from sources including Wikidata and ConceptNet \citep{speer2017conceptnet}. To mitigate the problem of information leakage from the head entity, \citet{poerner2019bert} propose LAMA-UHN, which is a hard subset of LAMA that has less token overlaps in head and tail entities. X-FACTR~\citep{jiang2020x} and mLAMA~\citep{kassner2021multilingual} extend knowledge probing to the multilingual scenario and introduce multi-token answers. They each propose decoding methods that generate multi-token answers, which we have shown to work poorly on \data. BioLAMA \citep{sung2021language} is a concurrent work that also releases a benchmark for biomedical knowledge probing. 

\stitle{Probing via Prompt Engineering.} Knowledge probing is sensitive to what prompt is used \citep{jiang2020can}. To bootstrap the probing performance, \citet{jiang2020can} mine more prompts and ensemble them during inference. Later works parameterised the prompts and made them trainable \citep{shin2020eliciting,fichtel2021prompt,qin2021learning}.
We have opted out prompt-engineering methods that require training data in this work, as tuning the prompts are essentially tuning an additional (parameterised) model on top of PLMs. As pointed out by \citet{fichtel2021prompt}, prompt tuning requires large amounts of training data from the task. Since task training data is used, the additional model parameters are exposed to the target data distribution and can solve the set set by overfitting to such biases \citep{cao-etal-2021-knowledgeable}. In our work, by adaptively finetuning the model with a small set of raw sentences, we elicit the knowledge out from PLMs but do not expose the data biases from the benchmark (\data).

\stitle{Biomedical Knowledge Probing.} 
\citet{nadkarni2021scientific} train PLMs as KB completion models and test on the same task to understand how much knowledge is in biomedical PLMs. BioLAMA focuses on the continuous prompt learning method OptiPrompt \citep{zhong2021factual}, which also requires ground-truth training data from the task. Overall, compared to BioLAMA, we have provided a more comprehensive set of probing experiments and analysis, including proposing a novel probing technique and providing human evaluations of model predictions.

%% file: sec-conclusion.tex
In this work, we created a carefully curated biomedical probing benchmark, \data, from the UMLS knowledge graph. We illustrated that state-of-the-art probing techniques and biomedical pre-trained languages models (PLMs) struggle to cope with the challenging nature (e.g. multi-token answers) of this specialised domain, reaching only an underwhelming $3\%$ of acc@10. To reduce the gap, we further proposed a novel contrastive recipe which rewires the underlying PLMs without using any probing-specific data and illustrated that with a lightweight pre-training their accuracies could be pushed to $24\%$.

Our experiments also revealed that different layers of transformers  encode different types of information, reflected by their individual success at handling certain types of prompts. Additionally, using a human expert, we showed that the existing evaluation criteria could overpenalise the models as many valid responses that PLMs produce are not in the ground truth UMLS knowledge graph. This further highlights the importance of having a human in the loop to better understand the potentials and limitations of PLMs in encoding domain specific factual knowledge.

Our findings indicate that the real lower bound on the amount of factual knowledge encoded by PLMs is higher than we estimated, since such bound can be continuously improved by  optimising both the encoding space (e.g. using our self-supervised contrastive learning technique) and the input space (e.g. using the prompt optimising techniques~\cite{shin2020autoprompt,qin2021learning}). We leave further exploration of integrating the two possibilities to future work.

%% file: sec-appendix.tex
\clearpage
\appendix
\section{Appendix}
\label{sec:appendix}
\subsection{Details of the Hardness Metrics}
In this paper, we use two automatic metrics to distinguish hard and easy queries. In particular, we first filter out easy queries by an exact matching metric (i.e. the exactly matching all the words of  answer from queries). Since our \data contains multiple answers for queries, we use a threshold on the average exact matching score, i.e. \textbf{avg-match}>0.1, to filter out easy examples, where \textbf{avg-match} is calculated by:
\begin{align}
  \textbf{avg-match} = \frac{\textbf{Count}(\text{matched answers})}{\textbf{Count}(\text{total answers})}.  \nonumber
\end{align}
This metric can remove all the queries that match the whole string of answers. However, some common sub-strings between queries and answers  also prone to reveal answers, particularly benefiting those retrieval-based probing approaches. E.g. <\textsl{Magnesium Chloride}, \rel{may-prevent}, \textsl{Magnesium Deficiency}>.
Therefore, we further calculate the $\textbf{ROUGE-L}$ score~\cite{lin-och-2004-automatic} for all the queries by regarding <query, answers> pairs as the <hypothesis, reference> pairs, and further filter out the $\textbf{ROUGE-L}$>0.1 queries.

\subsection{Hyperparameters Tuning} 
We train our \model based on 10k sentences which are randomly sampled from the original pre-training corpora of the corresponding PLMs. Since most of the biomedical BERTs use PubMed texts as their pre-training corpora, for all biomedical PLMs we sampled random sentences from a version of PubMed corpus used by BlueBERT model~\cite{peng2019transfer}, while for BERT we sampled sentences from its original Wikitext corpora. For the hyperparamters of our \model,  Table \ref{Table:search_space} lists our search options and the best parameters used in our paper. 

\begin{table*}[!ht] %
\small
\centering
\resizebox{.68\textwidth}{!}{
\begin{tabular}{lr}
\toprule
\textbf{Hyperparameters} & \textbf{Search space} \\
\midrule
rewire training learning rate & \{\texttt{1e-5}, \texttt{2e-5}$^\ast$,  \texttt{5e-5}\} \\
rewire training steps & 500 \\
rewire training mask ratio & \{0.1, 0.2, 0.3, 0.4$^\ast$, 0.5$^\ast$\} \\
$\tau$ in InfoNCE of rewire training & \{0.02,0.03$^\ast$,0.04,0.05\} \\
rewire training data size & \{1k, 10k$^\ast$, 20k,100k\} \\
step of checkpoint for probing & \{50, 150, 200$^\ast$, 250, 300, 350\} \\
\texttt{max\_seq\_length} of tokeniser for queries & 50 \\
\texttt{max\_seq\_length} of tokeniser for answers & 25 \\
\bottomrule
\end{tabular}
}
\vspace{-0.5em}
\caption{Hyperparameters along with their search grid. $\ast$ marks the values used to obtain the reported results.}
\label{Table:search_space}
\vspace{-1em}
\end{table*}

\subsection{The Impact of Mask Ratios}
\begin{figure}[htp]
    \centering
    \includegraphics[width=0.23\textwidth]{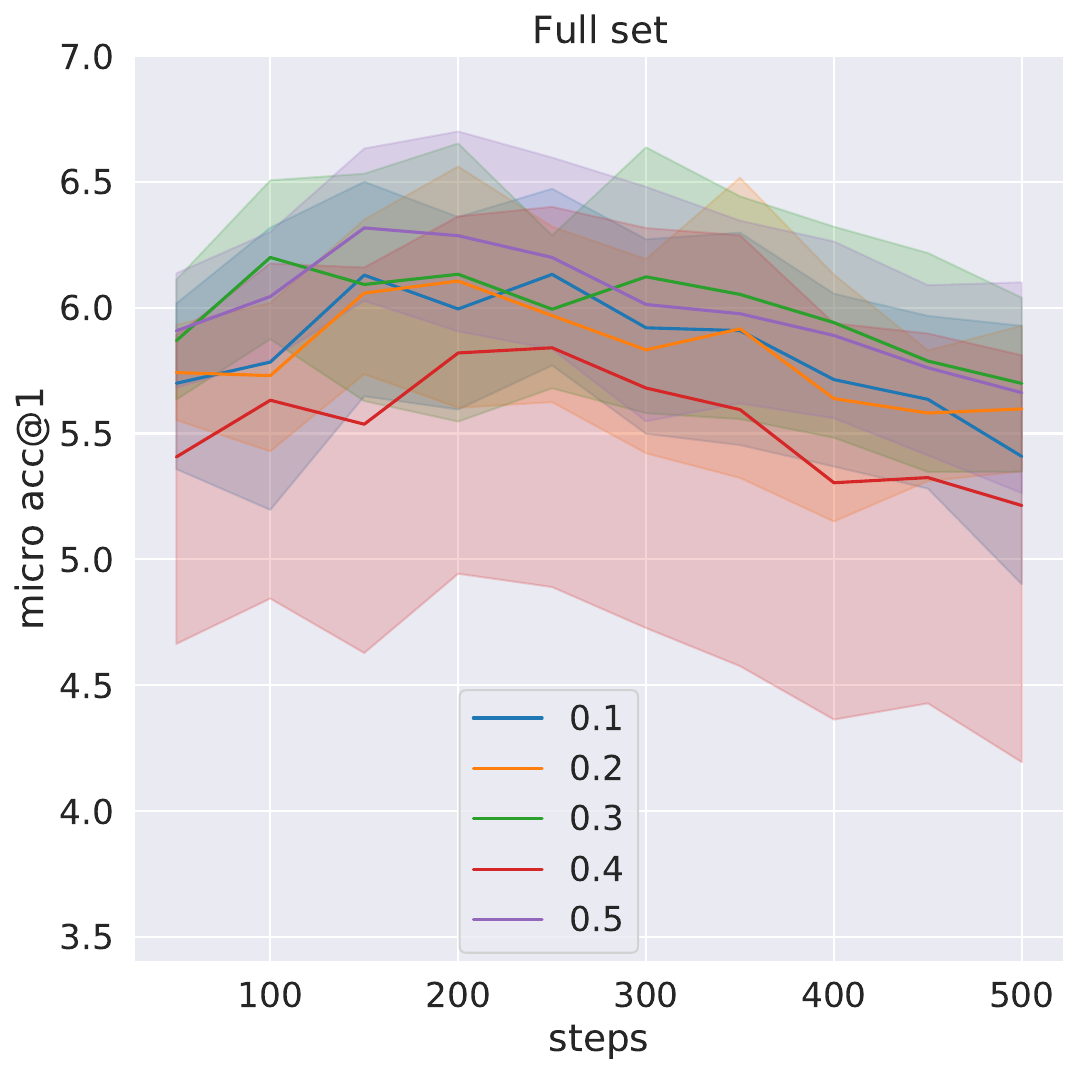}
     \includegraphics[width=0.23\textwidth]{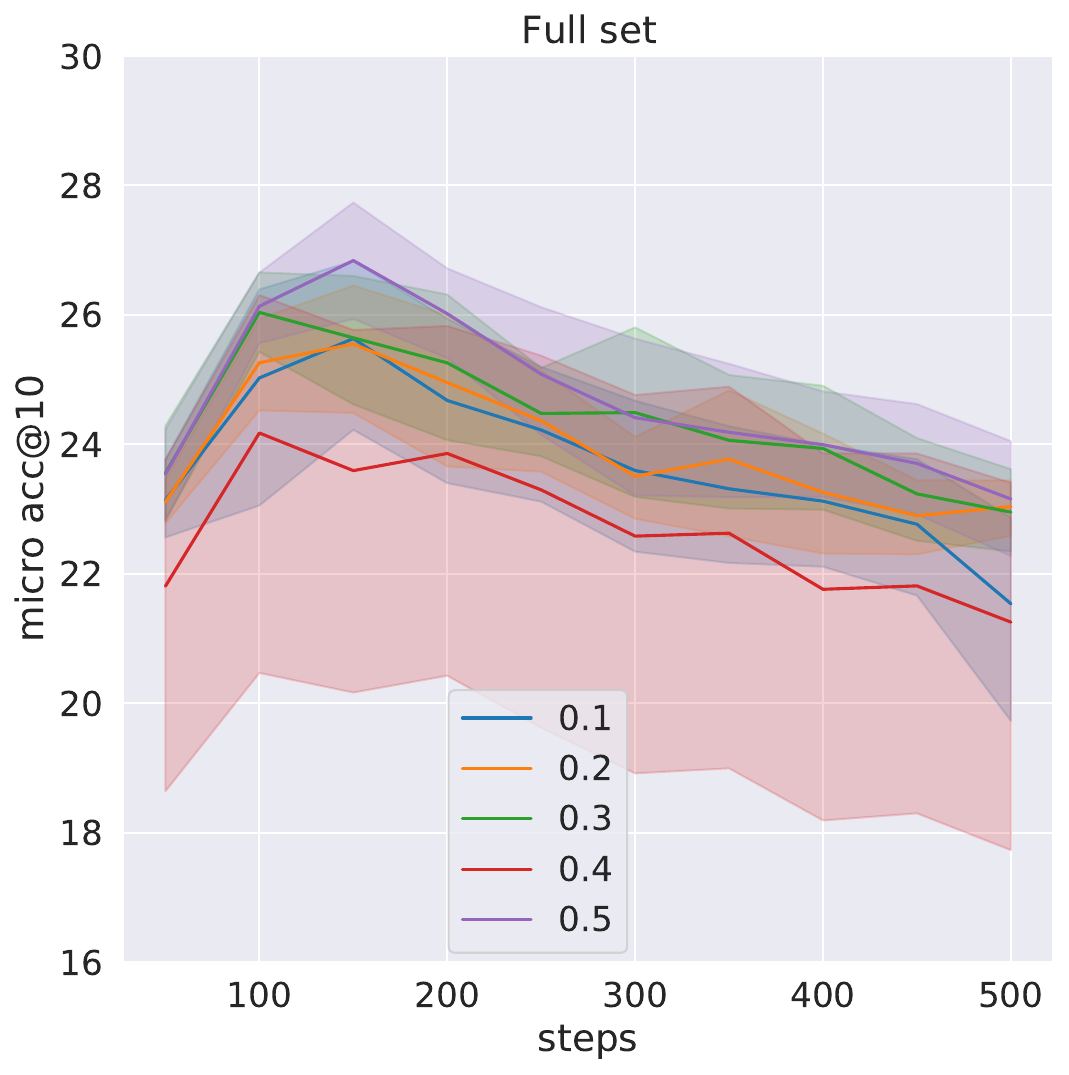}\\
    \includegraphics[width=0.23\textwidth]{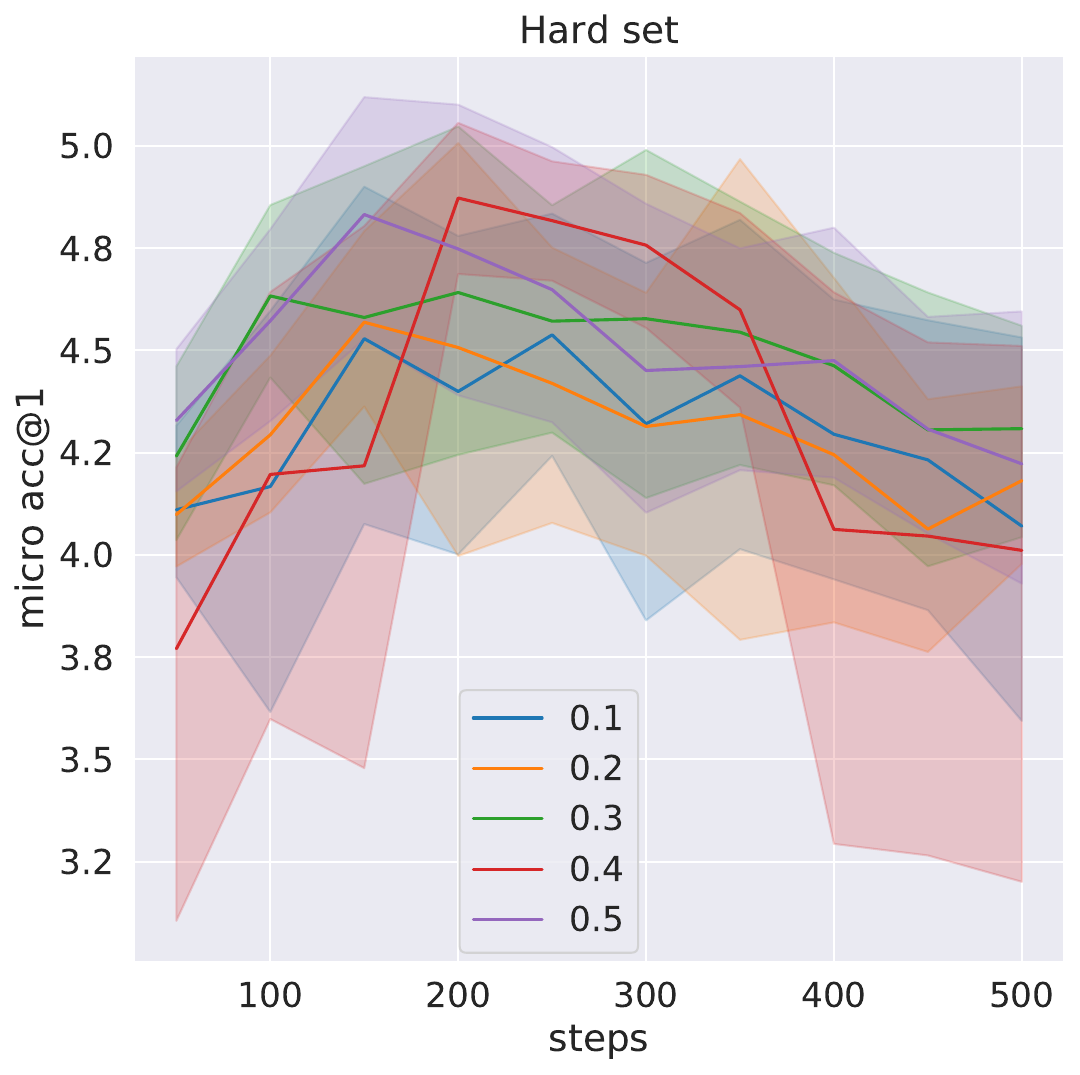}
     \includegraphics[width=0.23\textwidth]{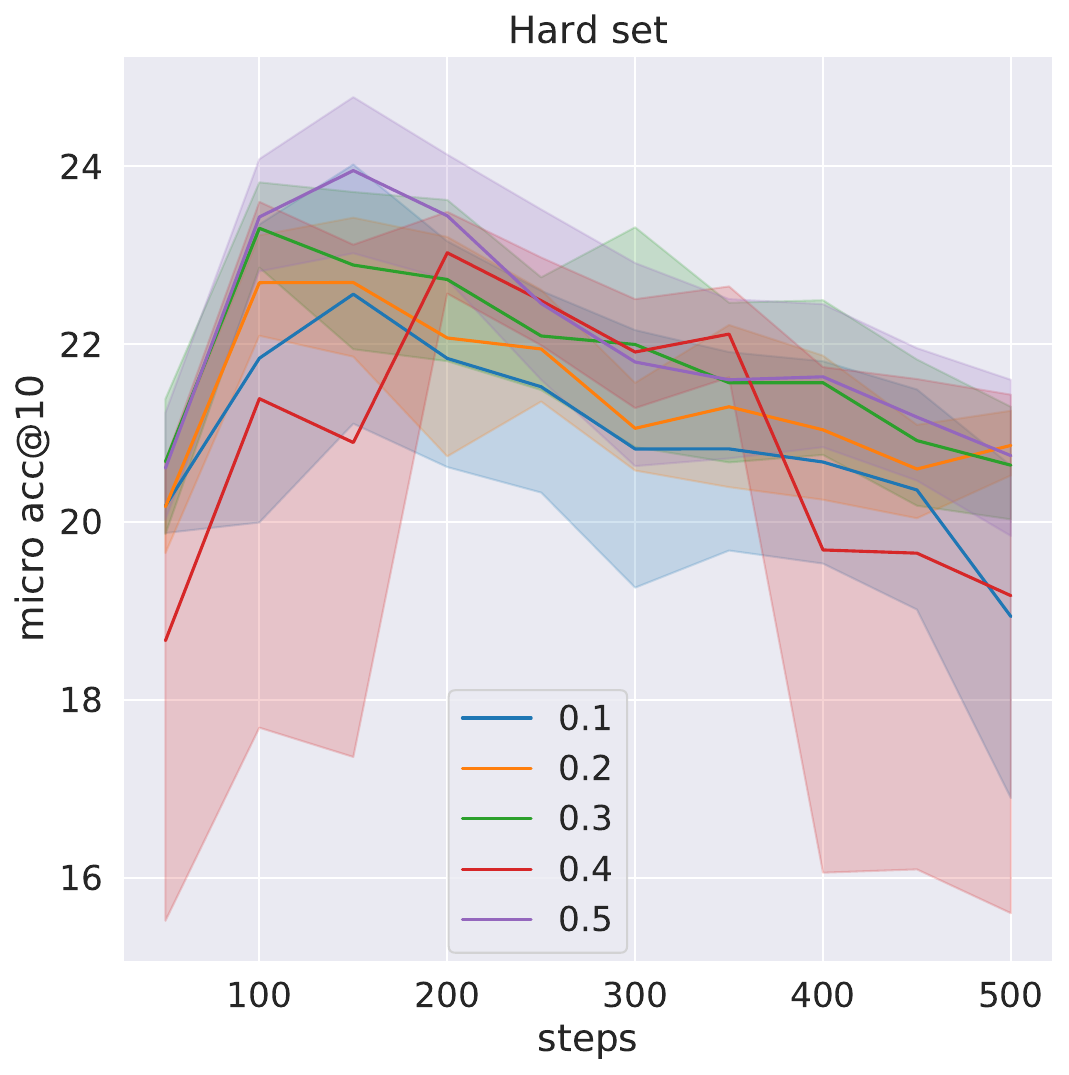}
     \vspace{-0.5em}
    \caption{Performance of \model based on PubMedBERT over different mask ratios. The shaded regions are the standard deviations under 10 different random sentence sets sampled from the PubMed corpus.}
    \label{fig:by_mask_ratios_step}
\end{figure}

To further investigate the impact of the mask ratio to the probing performance, we also test our \model based on PubMedBERT over different mask ratios (\{0.1, 0.2, 0.3, 0.4, 0.5\}) under the 10 random sentence sets, the result of which is shown in Figure \ref{fig:by_mask_ratios_step}. We can see that over different mask ratios the \model always reaches their best performance under certain pre-training steps. And the performance curves of mask ratios are different over the full and hard sets, but they all achieves a generally good performance when the mask ratio is 0.5, which validates that different mask ratios favour different types queries.

\begin{table*}[!ht]
    \centering
    \resizebox{.99\textwidth}{!}{
    \begin{tabular}{lllll}
    \toprule
    \multicolumn{5}{l}{\textbf{Query 1:} The gene product \underline{\entb{HLA Class II Histocompatibility Antigen, DP(W4) Beta Chain}} is encoded by gene \underline{[Y]} .} \\
    \multicolumn{5}{l}{\textbf{UMLS Answers:} \ent{MHC Class II Gene}, \ent{HLA-DPB1 Gene}, \ent{Immunoprotein Gene}} \\
    \multicolumn{5}{l}{\textbf{Human Answers:} \ent{MHC Class II Gene}, \ent{HLA-DPB1 Gene}} \\
    \midrule
    \textbf{Model} & \textbf{\model (PubMedBERT)} & \textbf{X-FACTR (BlueBERT)}  & \textbf{Generative PLMs (SciFive-large)} \\
    \multirow{5}{*}{\textbf{Top-5}} & \ent{MHC Class II Gene} & \entb{b} & \ent{HLA-DRB1} & \\
    & \entb{MHC Class I Gene} & \entb{hla} & \entb{encoding HLA} \\
    & \entb{HLA-A Gene} & \entb{dqb1} & \entb{DP(W)} \\
    & \ent{HLA-DPB1 Gene} & \entb{locus dqb1} & \entb{HLA-B} \\
    & \entb{HLA-F Gene} & \entb{2 , dq beta 2} & \entb{HLA-DQ} \\
    \midrule
    \midrule
    \multicolumn{5}{l}{\textbf{Query 2:} The gene product \underline{\entb{Tuberin}} is encoded by gene \underline{[Y]} .} \\
    \multicolumn{5}{l}{\textbf{UMLS Answers:} \ent{TSC2 Gene}, \ent{Signaling Pathway Gene}} \\
    \multicolumn{5}{l}{\textbf{Human Answers:} \ent{TSC2 Gene}, \ent{ Tuberin}} \\
    \midrule
    \textbf{Model} & \textbf{\model (PubMedBERT)} & \textbf{X-FACTR (BlueBERT)}  & \textbf{Generative PLMs (SciFive-large)} \\
    \multirow{5}{*}{\textbf{Top-5}} & \ent{TSC2 Gene} & \entb{family of tuberins} & \entb{``''} \\
    & \entb{SKA2 Gene} & \entb{\#\#t1} & \entb{TUB} \\
    & \entb{TSPY1 Gene} & \entb{symbol tuber} & \ent{Tuberin} \\
    & \ent{Tuberin} & \entb{( tuber )} & \entb{TUBE} \\
    & \entb{TSC1 Gene} & \entb{a} & \entb{TUBB} \\
    \midrule
    \midrule
    \multicolumn{5}{l}{\textbf{Query 3:} \underline{\entb{Refractory Monomorphic Post-Transplant Lymphoproliferative Disorder}} may have \underline{[Y]} .} \\
    \multicolumn{5}{l}{\textbf{UMLS Answers:} \ent{Lymphadenopathy}, \ent{Aggressive Clinical Course}, \ent{Extranodal Disease}} \\
    \multicolumn{5}{l}{\textbf{Human Answers:} \ent{Early post-transplant lymphoproliferative disorder}, \ent{ Lymphoproliferative disorder following transplantation }, } \\
    \multicolumn{5}{l}{\ent{ Refractory Polymorphic Post-Transplant Lymphoproliferative Disorder}, \ent{Aggressive Clinical Course}, \ent{Post transplant lymphoproliferative disorder}} \\
    \multicolumn{5}{l}{\ent{Neoplastic Post-Transplant Lymphoproliferative Disorder}, \ent{Refractory Monomorphic Post-Transplant Lymphoproliferative Disorder}} \\
    \midrule
    \textbf{Model} & \textbf{\model (PubMedBERT)} & \textbf{X-FACTR (BlueBERT)}  & \textbf{Generative PLMs (SciFive-large)} \\
    \multirow{5}{*}{\textbf{Top-5}}
    & \ent{Early post-transplant lymphoproliferative disorder} & \entb{manifestations} & \entb{similar to this} \\
    & \ent{Lymphoproliferative disorder following transplantation} & \entb{relapses} & \entb{in this study} \\
    & \ent{Refractory Polymorphic Post-Transplant Lymphoproliferative Disorder} & \entb{phenotype} & \entb{similar to our case} \\
    & \ent{Aggressive Clinical Course} & \entb{- specific phenotype} & \entb{similar to ours} \\
    & \ent{Post transplant lymphoproliferative disorder} & \entb{features} & \entb{similar to this case} \\
    \midrule
    \midrule
    \multicolumn{5}{l}{\textbf{Query 4:} \underline{\entb{moexipril}} might treat \underline{[Y]} .} \\
    \multicolumn{5}{l}{\textbf{UMLS Answers:} \ent{Diabetic Nephropathies}, \ent{Heart Failure}, \ent{Hypertension}, \ent{Ventricular Dysfunction, Left}} \\
    \multicolumn{5}{l}{\textbf{Human Answers:} \ent{Essential Hypertension}, \ent{ Hypertension}} \\
    \midrule
    \textbf{Model} & \textbf{\model (PubMedBERT)} & \textbf{X-FACTR (BlueBERT)} & \textbf{Generative PLMs (SciFive-large)} \\
    \multirow{5}{*}{\textbf{Top-5}} & \ent{Essential Hypertension} & \ent{hypertension} & \entb{``''}\\
    & \entb{Posttransplant hyperlipidemia} & \entb{diabetes mellitus} & \entb{this}\\
    & \ent{Hypertension} & \ent{essential hypertension} & \entb{them}\\
    & \entb{Atherosclerotic Cardiovascular Disease} & \entb{diabetes}  & \entb{migraine}\\
    & \entb{Type 1 Diabetes Mellitus} & \entb{in patients with hypertension}  & \entb{patients}\\
    \bottomrule
    \end{tabular}
    }
    \caption{Example predictions of different probing approaches.  The human answers are annotated based on the \model predictions.}
    \label{tab:predictions}
\end{table*}